# Precision Cotton Disease Management: Leveraging UAV Imagery and Custom YOLO Deep Learning Framework for Enhanced Detection

Anwar Iqbal*[1] aiqbal.rime23smme@student.nust.edu.pk

Kunwar Faraz Ahmad Khan[1] kunwar.faraz@smme.nust.edu.pk

Shahbaz Khan[1] shahbaz.khan@smme.nust.edu.pk

Khawaja Fahad Iqbal[1] fahad.iqbal@smme.nust.edu.pk

Sara Ali[1,2] sara.baber@smme.nust.edu.pk

Saqib Nazir[1] saqib.nazir@smme.nust.edu.pk

Muhammad Azhar Javed[3] muhammadazhar6231@gmail.com

1 Department of Robotics and Artificial Intelligence, SMME National University of Science and Technology (NUST) Islamabad, Pakistan

2 Suzhou Industrial Park Monash Research Institute of Science and Technology, Suzhou, PR China

3 Department of Plant Pathology Bahauddin Zakriya University Multan, Pakistan

## Abstract:

*Cotton sustains the textile economies of over 80 countries with an annual yield of 113 million bales, yet 30–60 % yield losses are still attributed to viral, bacterial and fungal diseases that remain undetected until manual field scouting. To address this critical issue, the use of Unmanned Aerial Vehicle (UAV) imagery combined with deep learning offers a scalable early-warning system, but existing models suffer from low early-stage accuracy, poor transferability and high computational cost. In this paper, we propose Cotton-Vision, a fully convolutional single-stage detector that synergizes three light-weight architectural innovations. The architectural changes include ghost bottleneck modules for 20 % parameter reduction, efficient channel-attention (ECA) for adaptive feature recalibration, and cross-stage partial (CSP) fusion for enriched gradient flow. The network is trained on a local dataset of cotton collected from 7 different fields and comprising over 2700 drone images from local cotton farms in the Punjab Province of Pakistan. Extensive augmentation (rotation, shear, exposure, saturation) yields 11704 training samples. On a held-out test set, Cotton-vision achieved a precision of 97.7%, a recall of 77.7%, an mAP@50 of 85.6%, and an mAP@50-95 of 76.6%. These results represent a 10-20% improvement in precision, recall and mAP over the leading models, including YOLOv5, YOLOv8, YOLOv11, YOLOv12, RT-DETR and RF-DTR. Through a comprehensive ablation study, we demonstrate that each of our proposed architectural enhancements contributes to the model's superior performance with 12%-16% improvement from the base model. Our findings confirm that this synergy of lightweight and attention-based modules provides a robust and computationally viable solution for early cotton disease detection.*

## 1. Introduction:

Cotton provides 31 % of global textile fiber and supports 250 M livelihoods [1]. Pakistan, the fifth-largest producer with 8.3 million bales of cotton production, relies on cotton for 8.2% of GDP and 55% of export earnings [2]. However, total 40–60% of total annual yield losses are attributed to biotic stresses, predominantly Cotton Leaf Curl Virus (CLCuV), bacterial blight, Verticillium wilt, Alternaria leaf spot and the boll-worm complex [3]-[4]. Early symptoms are cryptic and heterogeneous, often confused with nutrient deficiency or drought stress. Delayed intervention results in insect-vector proliferation and secondary infections, amplifying economic damage[5].

The traditional methods of visual inspection and scouting cover less than 5% of the planted area, however the process is labor-intensive [6][7] . The reliance on naked-eye assessment by agronomists introduces significant limitations, including the requirement for continuous expert monitoring, substantial costs, and the challenge of accurately identifying diseases at early stages when intervention would be most effective [8]. Manual approaches are particularly inadequate for vast agricultural operations and remote farming areas, where timely expert consultation may not be readily available [9]. Furthermore, the subjective nature of visual assessment creates inconsistencies in disease diagnosis, potentially leading to inappropriate treatment decisions and resource misallocation [10].

The emergence of precision agriculture technologies, particularly the integration of Unmanned Aerial Vehicle (UAV) imagery with advanced deep learning frameworks, presents transformative solutions to these challenges [11]. UAV-based proximal sensing offers centimeter-resolution imagery with rapid revisit capabilities, while deep convolutional networks automatically learn hierarchical disease signatures from high-dimensional image data [12]-[13]. This technological convergence enables the development of scalable early-warning systems that can monitor extensive cotton fields with unprecedented accuracy and efficiency [14]. Recent advances in computer vision and artificial intelligence have demonstrated the potential to revolutionize crop health monitoring by providing detailed assessment capabilities and facilitating data-driven decision-making processes [15-16] .

Despite significant progress in agricultural deep learning applications, existing models face critical limitations that hinder their practical deployment in cotton disease management. Many current approaches suffer from computational inefficiency, making them unsuitable for processing on crops of remote areas with limited data and computational resources [17]. The complexity of cotton disease manifestations, including varied symptom presentations, co-occurrence of multiple diseases, and diverse field conditions, necessitate robust and adaptable detection systems [18].

The development of lightweight yet accurate deep learning architectures specifically tailored for cotton disease detection represents a critical research imperative [19]. Modern agricultural applications demand models that can achieve high detection accuracy while maintaining computational efficiency suitable for deployment on resource-constrained computational platforms [20]. This technological need aligns with the broader goals of sustainable agriculture, where precision application of interventions can significantly reduce chemical inputs, minimize environmental impact, and optimize resource utilization [21]. The integration of advanced attention mechanisms, efficient feature extraction techniques, and optimized neural network architecture offers promising pathways to overcome current limitations while advancing the state-of-the-art in agricultural monitoring systems [8][21]-[22].

Our work addresses these limitations by introducing Cotton-Vision, a novel deep learning framework designed to be both highly accurate and computationally efficient. The core contributions of this paper are:

1. The development of Cotton-Vision, a lightweight object detection model incorporating Ghost Bottleneck, ECA, and CSP modules for optimized performance.
2. The creation and release of Sanjrani_CottonDB, a new, high-resolution UAV-RGB dataset specifically for detection of cotton disease, which includes diverse imagery of cotton at various heights.
3. A comprehensive comparative analysis demonstrates that Cotton-Vision significantly outperforms state-of-the-art CNN models on a held-out test set.
4. An ablation study confirming the 12% to 16% increase in precision of each architectural innovation within our proposed framework.

The paper is organized as follows: Section 1 introduces the significance of precision cotton disease management. Section 2 reviews related work on UAV imagery and deep learning models. Section 3 details data collection, dataset generation, and the proposed Cotton-Vision framework. Section 4 describes experimental setup and training strategies. Section 5 presents results and evaluations, followed by conclusions in the final section.

# 2. Related Work

The application of machine learning (ML) and deep learning (DL) methodologies, frequently integrated with Unmanned Aerial Vehicles (UAVs) for remote sensing, has become fundamental to advancing precision agriculture [13][23][24]. Recent scientific studies prioritize the development of advanced models that exhibit enhanced accuracy, computational efficiency, and superior generalization capabilities when deployed in complex, real-world agricultural settings [25]. This section reviews several high-performing models and innovative methodologies identified in the literature, focusing specifically on cotton disease identification, UAV-based detection systems, and lightweight architectures suitable for informed pest management in agriculture.

## 2.1YOLO Family for Agriculture

The You Only Look Once (YOLO) family has emerged as the cornerstone of real-time object detection in agricultural applications, demonstrating exceptional performance in UAV-based imagery processing [26]. The evolution from YOLOv1 to the latest YOLOv12 represents a continuous advancement in balancing detection accuracy with computational efficiency, making these architectures particularly suitable for precision agriculture deployment [27]. YOLOv5 established itself as a benchmark for agricultural applications due to its ease of use and computational efficiency [28]. Liu et al. [29] demonstrated YOLOv5's effectiveness for aerial cotton disease detection, achieving 99.2% detection on cotton seeds. Xu et al. [30] introduced YOLOv8 with significant architectural improvements including anchor-free detection heads and enhanced feature pyramid networks. The model's versatility in supporting multiple tasks object detection, instance segmentation, and pose estimation make it particularly valuable for comprehensive crop monitoring systems but it struggles in channel-based feature extractions.

Zhang et al. [31] proposed an improved YOLOX model incorporating Efficient Channel Attention (ECA) modules, hard-Swish activation functions, and Focal Loss for detecting cotton diseases and pests in complex backgrounds. This model achieved a mean Average Precision (mAP) of 94.60% with an inference speed of 74.21 FPS, significantly outperforming classical detection algorithms including Faster R-CNN (83.10% mAP) and baseline YOLOv5 (86.27% mAP). Yet, this model didn't provide a breakthrough in UAV based imagery to achieve highest accuracy of 94.60%. YOLOv11 represents the latest evolution in the YOLO family, incorporating Cross-Stage Partial with Self-Attention (C2PSA) modules and replacing C2f blocks with more efficient C3k2 blocks. The model's enhanced spatial attention mechanisms improve detection accuracy for small and overlapping objects, addressing key challenges in dense canopy environments typical of cotton fields.

Noon et al. [32] developed an enhanced YOLOX model specifically for handling multiple cotton plant diseases, achieving 93.4% accuracy in simulating different disease severity levels while simultaneously classifying co-occurring symptoms. Their approach incorporated a novel loss function to balance detection of primary and secondary disease manifestations, addressing a critical challenge in multi-disease scenarios. The progression from YOLOv5 to YOLOv12 demonstrates consistent improvements in parameter efficiency and inference speed. The latest YOLO26 eliminated Non-Maximum Suppression (NMS) requirements through dual assignment strategies, while YOLOv12 introduced Area Attention (A2) modules and Residual Efficient Layer Aggregation Networks (R-ELAN) for enhanced performance. These architectural advances collectively address the computational constraints inherent in precision agriculture applications where scare data processing on resource-limited hardware is essential. These advancements in architecture point out the need for innovative design by making the application of architectures more convenient and robust.

## 2.2 Lightweight and Attention-Based Models

Addressing the computational constraints inherent in UAV-based systems, researchers have increasingly focused on lightweight architectures that maintain detection accuracy while reducing computational overhead. Quan et al. [21] proposed a lightweight model for real-time field disease identification, achieving 89.3% accuracy with a processing time of just 0.3 seconds per image through novel pruning and quantization methods. Their approach demonstrated that computational efficiency and detection accuracy could be balanced through careful architectural optimization. Rai and Pahuja [33] developed an improved deep convolutional neural network for diseased cotton leaf classification, utilizing spatial attention mechanisms to reach 96.8% accuracy across seven disease categories. The model's strength lies in its ability to maintain high performance even with limited training data through suitable data augmentation techniques. However, these approaches primarily focused on handheld device deployment and had not been extensively tested for varying UAV flight heights and ground sample distances.

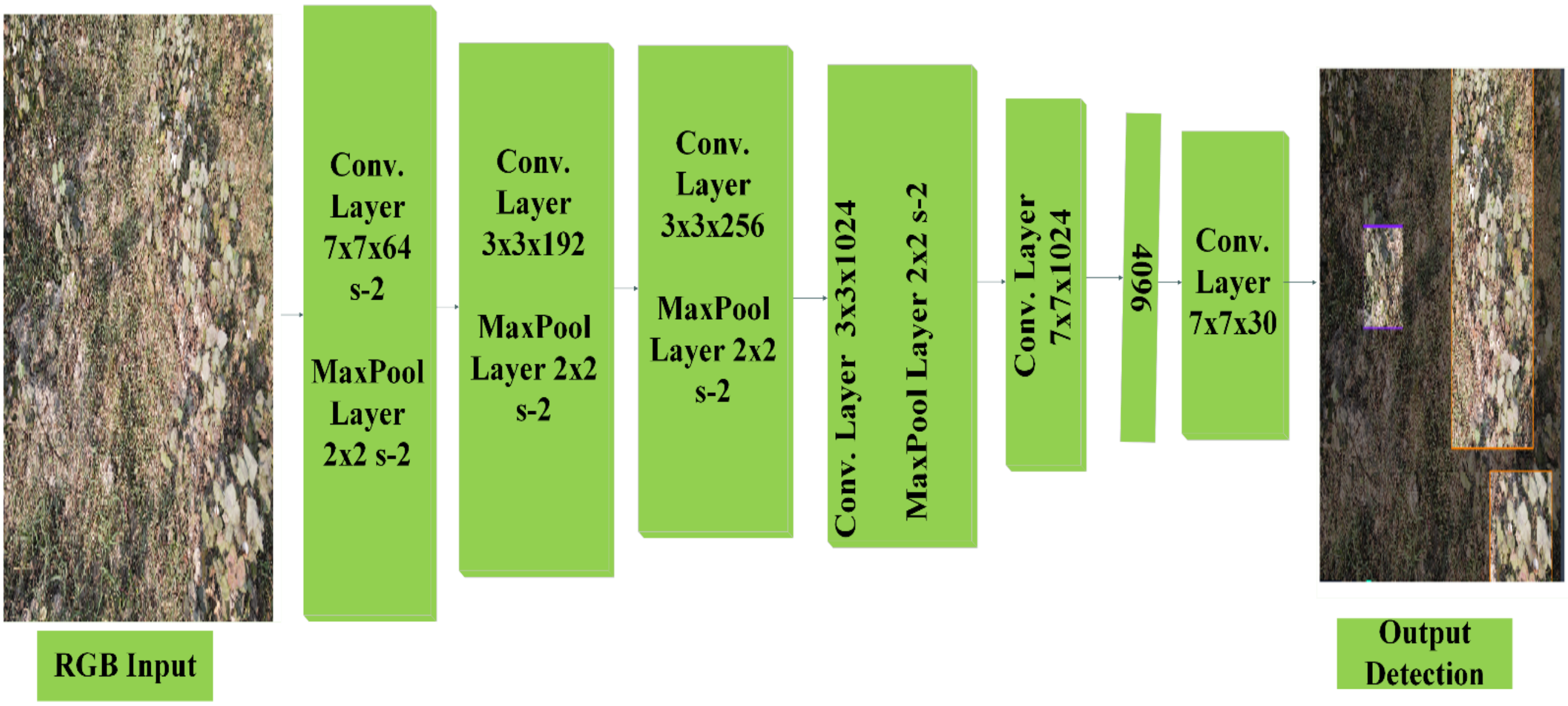

***Figure 1** YOLO architecture schematic diagram for Disease Detection in cotton field*

**Figure 1** schematically explains that YOLO (You Only Look Once) is a single-stage object detection architecture that performs object localization and classification in a single forward pass through a convolutional neural network. It divides the input image into a grid, where each cell predicts bounding boxes along with confidence scores and class probabilities, enabling fast and efficient real-time detection. **Table 1** in the same manner explains various deep learning methods employed for UAV Imagery to evaluate the performance of various deep learning method.

***Table 1** Comparative Analysis of Deep Learning Models for Different Datasets*

| Reference | Deep Learning Tool | Accuracy Score | Dataset / Focus |
|---|---|---|---|
| [34] | Deep transfer learning (ResNet variants) | 94.5% | Aerial surveillance for cotton disease |
| [35] | Integrated CNN + transfer learning | 95.63% | Early cotton disease detection via UAV |
| [36] | Hybrid Swin Transformer + CNN | 98.1% | Cotton leaf disease datasets |
| [33] | Improved CNN (4 conv layers) | 98.75% | Diseased vs. healthy cotton leaves |
| [37] | Snapshot ensemble + Vision Transformers | 97.2% | Cotton disease datasets |
| [32] | Improved YOLOX (α-IoU loss) | 73.13% | Southern Punjab cotton leaf curl disease |
| [9] | YOLOv5 for aerial images | 93.8% | Cotton disease detection from UAV |
| [38] | GIS + deep learning | 95.8% | Volunteer cotton detection |
| [39] | ConvNeXt + attention mechanisms | 95.9% | Lab-conditioned cotton disease detection |

## 3. Materials and Methods

This section details the material and methods to conduct the study including data collection, dataset generation, the proposed deep learning model for cotton disease detection, and the methodology used for evaluation.

### 3.1Data Collection and Preparation:

To facilitate the precise identification and localization of cotton diseases, high-resolution (4k and 1080 remote sensing images were collected using unmanned aerial vehicles (UAVs). Data collection was performed in cotton fields in locality to ensure diversity in dataset. A data collection exercise was run in Jampur, Punjab, Pakistan, during the high incidence season of cotton leaf curl virus from mid-August to late September 2024. Images were collected under natural light conditions in different weather (sunny and cloudy days) and periods (between 1 PM and 6 PM). **Table 2** indicates the varieties and field ID of the study used for data collection. The diversity in growth stages and image count is critical for robust model training and testing, as disease manifestation varies significantly across the plant lifecycle. The first field variety is developed by Nuclear Institute for Agriculture and Biology (NIAB), which is a significant agricultural research institute in Pakistan. The second variety (BT-786) is transgenic cotton containing a gene from the bacterium *Bacillus thuringiensis*. This gene allows the cotton plant to produce an insecticidal protein that is toxic to certain lepidopteran pests, such as the cotton bollworm, providing inherent pest resistance. SS -32 and Z-33 are local varieties of cotton while Cyto-1 and Cyto-2023 are recently released varieties of Cyto-Genetics.

***Table** 2 Data Collection Fields along with GPS Locations and varying growth stages. The dataset includes healthy and diseased cotton leaf curl virus images through UAV Platforms.*

| Field No./Cotton Variety | No. of Images | Field GPS Location (Lat, Long) | Date | Growth Stage | Primary Disease | Weather Condition |
|---|---|---|---|---|---|---|
| **F1/NIAB-78** | 323 | 29.674835,70.574835 | 14 Aug 2024 | Flowering | Bacterial Blight | Sunny |
| **F2/BT-786** | 106 | 29.676122,70.574694 | 14 Aug 2024 | Flowering | CLCuV | Sunny |
| **F3/SS-32** | 211 | 29.675354,70.585057 | 14 Aug 2024 | Flowering | CLCuV | Sunny |
| **F4/Cyto-1** | 633 | 29.651195,70.565722 | 03 Sep 2024 | Boll Opening | CLCuV | Sunny |
| **F5/SS-32** | 369 | 29.663936,70.577625 | 03 Sep 2024 | Boll Opening | CLCuV | Sunny |
| **F6/Z-33** | 646 | 29.558161,70.521473 | 20 Sep 2024 | Boll Opening | CLCuV | Partially Cloudy |
| **F7/Cyto-2023** | 434 | 29.5473741,70.638105 | 20 Sep 2024 | Boll Opening | Nil | Sunny |

The dataset consists of images of healthy and Cotton Leaf Curl Virus (CLCuV) affected cotton plants, collected from seven unique fields. These fields vary in cotton varieties, growth stages. **Figure 2** shows the fields' location of present throughout the locality. In 7 fields of cotton, it can be noted that 5 varieties have been selected to study the impact of cotton leaf curl virus on given fields. The control group Cyto-2023 has no significant disease symptoms.

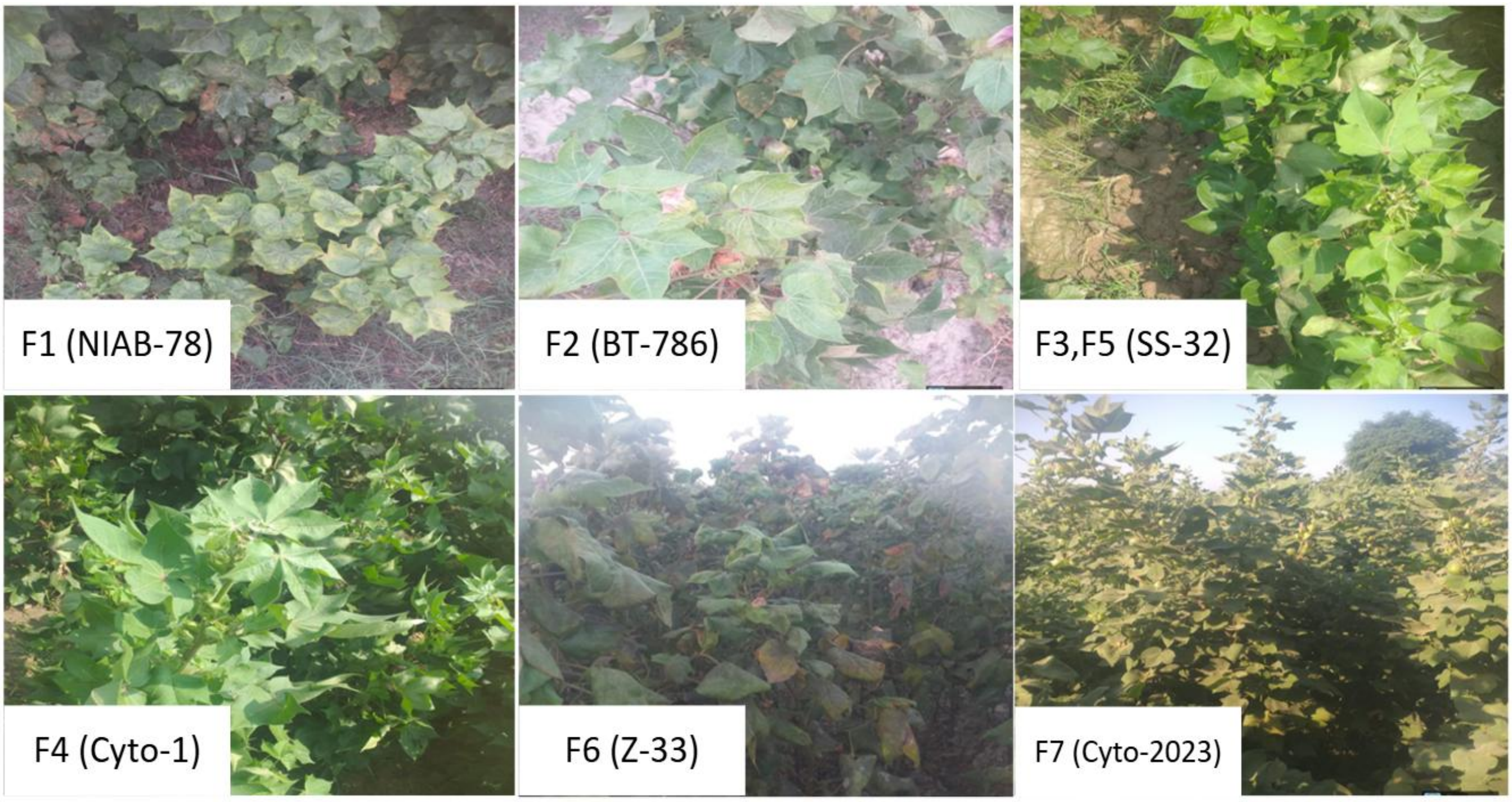

**Figure 2** Field Varieties in the Locality of Punjab, Pakistan

## 3.2 Flight Parameters

The UAVs and sensors used included a DJI Phantom 4 Pro V2 quadcopter equipped with a 20-megapixel digital RGB camera and a DJI Mavic Air 2 drone with 48-megapixel digital RGB. These drones were flown at an altitude of 2 m,5 m and 10 m above ground level (AGL) at a speed of 2.8 m/s. To minimize shadows, flights were conducted between 1:00 PM and 6:00 PM under natural light conditions. The resulting images have a resolution of 1920 × 1080 pixels and a Ground Sampling Distance (GSD) of 0.2 cm. Data collection was ideally carried out under clear and optimal weather conditions to avoid disturbances and ensure accuracy. The images are captured with 60% frontal and 60% side overlapping with camera angle directly pointing towards the ground.

## 3.3 Dataset Generation

The creation of an inclusive, robust dataset is a critical prerequisite for developing a reliable deep learning model for agricultural disease detection. The complexities such as varied symptom presentations, co-occurrence of multiple diseases, and diverse field conditions were handled with precise image selection for disease categories to make an inclusive dataset. Our dataset, called Sanjrani_CottonDB, was generated and annotated using the Roboflow platform, following a multi-step pipeline to ensure data quality and consistency.

### 3.3.1 Image Preprocessing

Image preprocessing is a critical step in preparing raw drone-captured data for deep learning models, ensuring consistency and enhancing feature learning. The process described involved four main steps within the Roboflow environment. First, Auto-orient corrected image rotation using metadata. Second, images were resized to a standard resolution to optimize batch processing. Third, converting images to grayscale mitigated potential color channel issues and simplified the data as model might get grey color of CLCuV as primary feature. Finally, tiling 2*2 broke high-resolution images into four smaller, manageable sections, improving both computational efficiency and the detection of small objects within the large cotton fields.

### 3.3.2 Data Annotation and Plant Pathologist Validation

The task of data annotation was paramount for training the deep learning model and was conducted with a high degree of precision. This method involves manually placing points densely along the contour of a diseased region to create a closed polygon, ensuring precise localization of symptoms rather than coarse bounding rectangles. To guarantee the accuracy and biological validity of these annotations, the process was overseen by qualified researchers of plant pathology. A stringent quality control protocol was implemented to maintain annotation integrity. This rigorous, expert-guided annotation process is a cornerstone of the dataset's quality, providing highly accurate labels that are essential for training a model capable of discerning subtle and early-stage disease symptoms. **Figure 3** demonstrates the annotation through bounding box methodology using Roboflow Platform. The area of bounding box corresponds to the disease localization where the disease is localized in the coordinates. The size variation is validated by plant pathologist. Expert opinions were solicited for validation purposes. All participating experts were fully informed about the study objectives and provided verbal/written informed consent prior to their participation. No personal or sensitive data was collected.

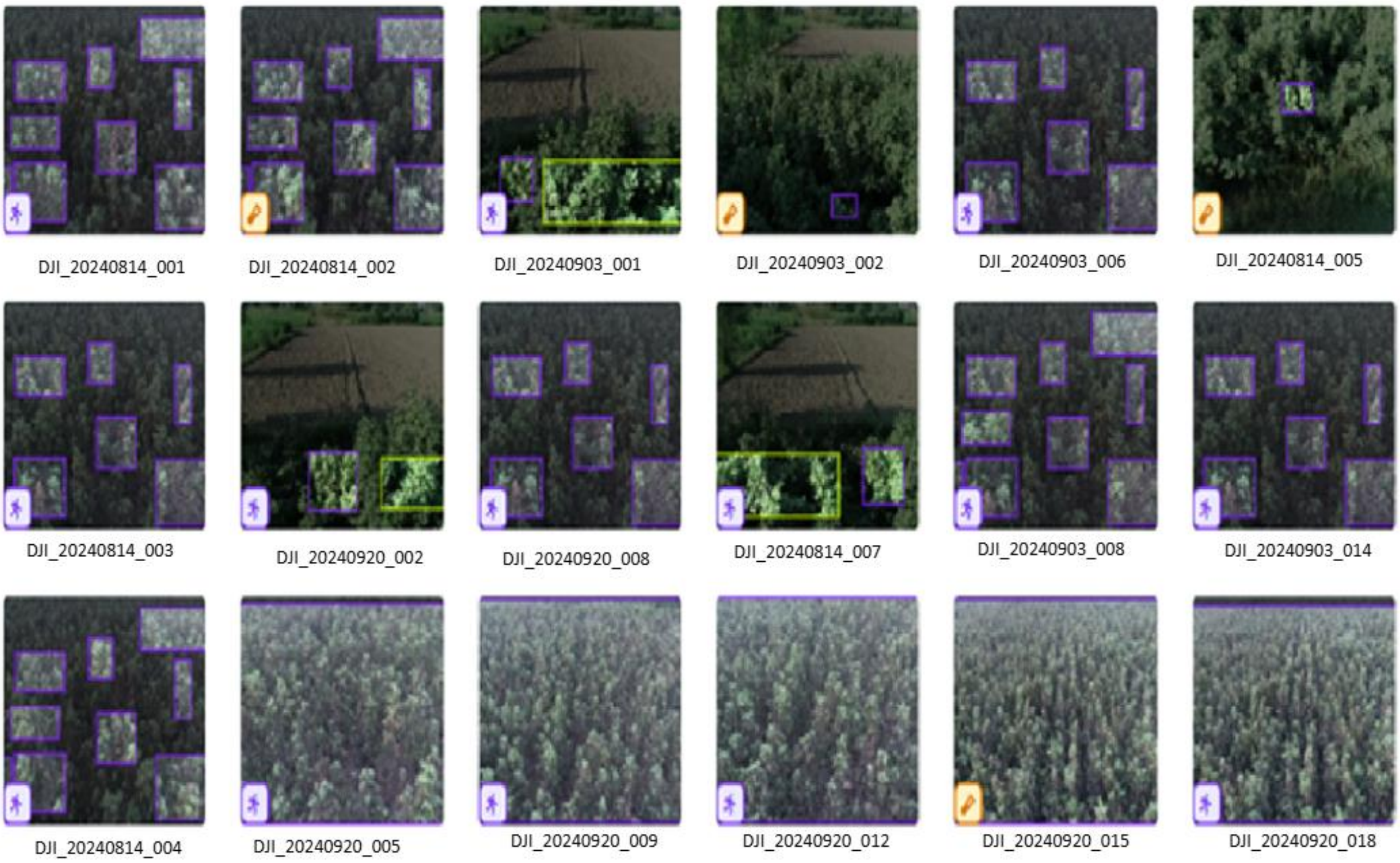

**Figure 3** Data Annotation of the captured images with image Identification number using Roboflow in consultation with Plant Pathologist. The images are divided into training and test with purple color representation of training and brown tag as testing.

## 3.5 Mathematical Approaches:

The mathematical framework underpinning Cotton-Vision is adopted to address the specific challenges of disease detection in UAV imagery. At its core, convolutional operations form the foundation for hierarchical feature extraction, enabling the model to learn discriminative patterns, from low-level textures and edges to high-level semantic concepts, directly from raw pixel data. The mathematical approach of convolution is given as.

a. $I_{(m,n)}$ is the input image,

b. $K_{(i,j)}$ is the kernel/filter.

In cotton disease detection context, this standard method is often modified to address specific challenges;

i. **Variable-sized receptive fields:** the studies by [31] [40]-[41] maintained that employing multiple kernel sizes (3×3, 5×5, and 7×7) in parallel branches significantly improved detection of varied symptom patterns in cotton leaves, achieving 94.5% accuracy compared to 91.2% with standard single-scale approaches.

ii. **Depth-wise Separable Convolution:** Quan et al. [42] applied depth-wise separable convolutions to reduce computational complexities while maintaining the performance.

$$S_{depth-wise}(i,j,k) = \sum_m \sum_n I(i+m, j+n, k) \cdot K_k(m,n) \quad (1)$$

$$S_{point-wise}(i,j,k) = \sum_{k'} S_{depth-wise}(i,j,k') \cdot P_k(k') \quad (2)$$

This adaptation reduced model size from 27M to 3.3M parameters while maintaining detection accuracy above 97%. This equation provides network layer wise operation for disease mapping.

### 3.5.1 Model Optimization and Performance:

a. Mean Average Precision Calculations:

$$mAP = \frac{1}{n}\sum_{i=o}^{n} AP_i \quad (3)$$

$$AP = \int_0^r p(r) \quad (4)$$

mAP and AP explaining

The Eq. 3,4 measure the overall performance of cotton disease detection systems across multiple disease classes where AP is the representation of each class calculated by integrating precision and recall curve as endorsed by Zhang et al. [43] and Wang et al. [44]. These equations provide the basis for disease detection using deep learning through UAV imagery, spanning preprocessing techniques, network architectures, performance matrices and optimization strategies to optimize the score for disease detection.

## 3.4 Proposed Model

The core of this research lies in the advanced deep learning approach designed to accurately identify and classify cotton diseases from drone imagery. **Figure 4** indicates complete process of vision pipeline. This process is critical to simulate the vast variability encountered in real-world UAV operations including changes in lighting, scale, and orientation, thereby building inherent capability into the model before training even began. Our proposed model is not only based on the YOLOv11 architecture but also incorporates three novel modifications to improve efficiency and performance. The process of ghost convolution where both global and local features are extracted with Sigmoid Loss function (SiLu). A Transformer based feature extraction module Efficient Channel Attention (ECA) is incorporated also to capture contextual information related to the position and distribution of diseased areas. Given that cotton plants are typically arranged in rows, and diseases might spread in specific patterns, understanding spatial distribution is crucial. The Transformer's self-attention and multi-head attention mechanisms help in focusing on all input positions simultaneously, capturing long-range dependencies and global contextual information, enhanced by positional encoding.

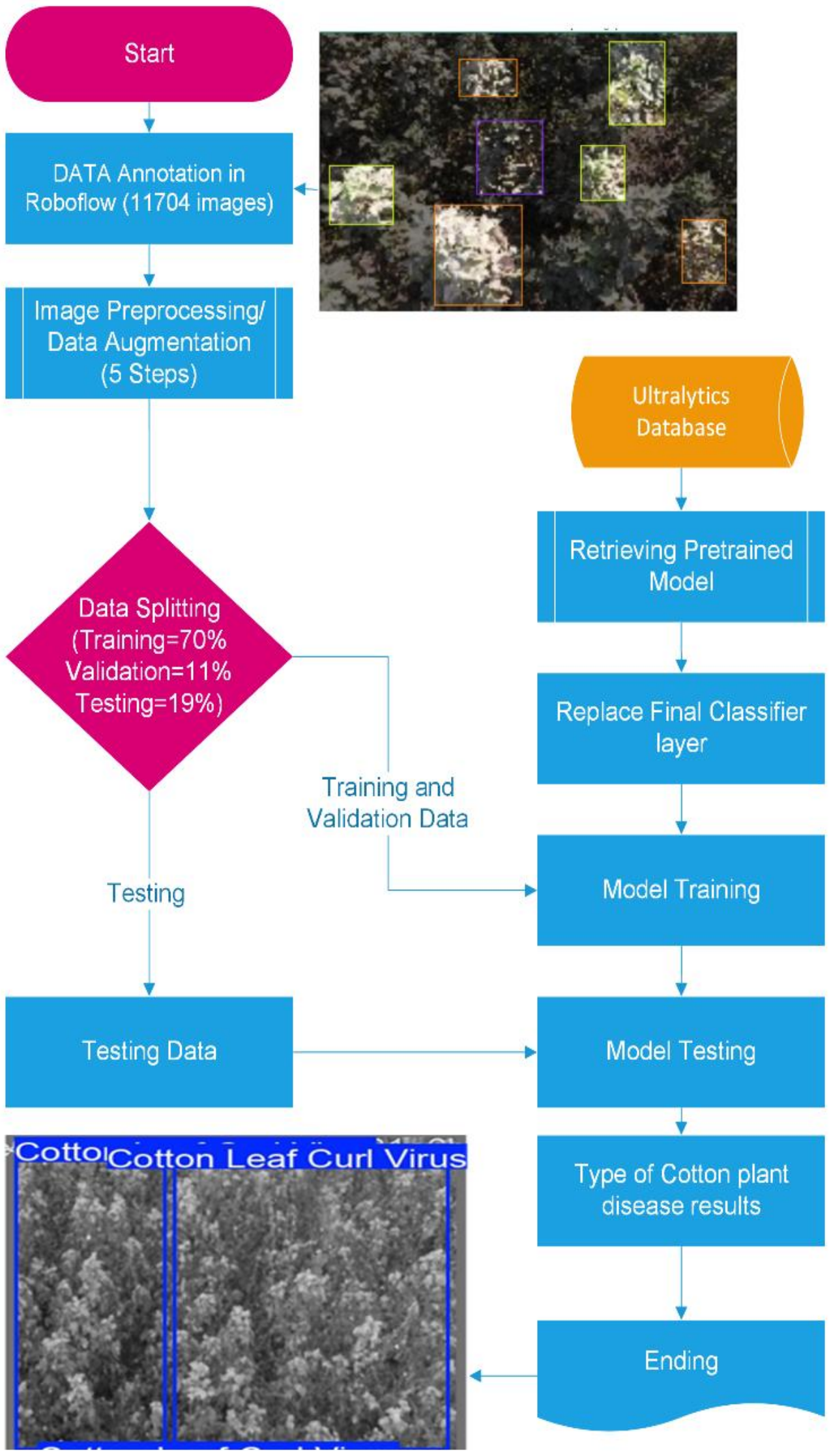

***Figure 4*** *Overall Implementation Pipeline of disease detection in cotton fields using deep learning through UAV Imagery*

### *A. Ghost Bottleneck*

Standard convolutional layers often generate feature maps with significant redundancy in the backbone. To address this inefficiency, we replace standard bottleneck blocks with Ghost Bottleneck modules [42]. **Figure 5** provides an illustration of the Ghost Convolution. The Ghost module generates a set of intrinsic feature maps through standard convolution and then uses cheap linear operations to create redundant 'ghost' feature maps, effectively expanding channel capacity and feature diversity of the network with a minimal computational cost [45]-[46]. The integration of Ghost Bottlenecks reduces the number of parameters and FLOPs by 20% compared to standard convolutions, making the model significantly lighter without a loss in representational power.

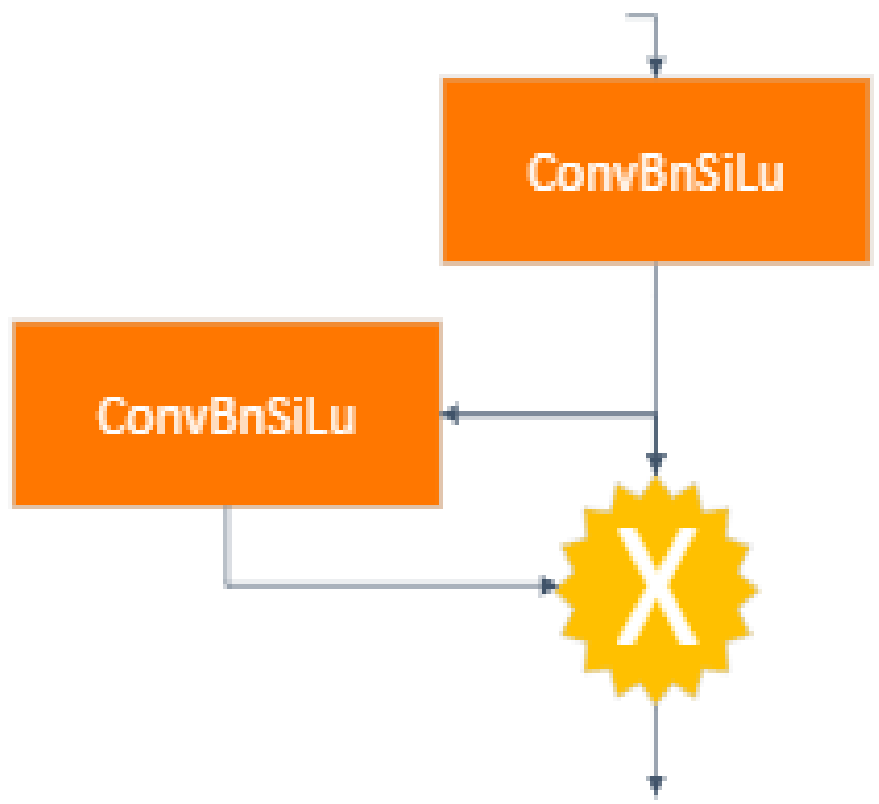

***Figure 5** Ghost Convolution Process for extracting the features*

### *B. Efficient Channel Attention (ECA)*

This adaptive feature refinement is crucial for discerning subtle, early-stage diseases. To enhance the model's focus on disease-specific features amidst complex field backgrounds, we integrate the Efficient Channel Attention (ECA) module into the backbone [47]. Unlike other attention mechanisms Global local attention or cross-channel attention that require dimensionality reduction, ECA avoids this overhead by performing a local cross-channel interaction. It uses a fast 1D convolution with an adaptive kernel size to compute channel attention weights as depicted in **figure 6**. This adoptive average pooling allows model weight and learning parameters adjustments. Which results in the model dynamically recalibrating feature maps, emphasizing informative channels related to disease symptoms and suppressing less useful ones. This selective emphasis improves the overall score of the model, leading to more accurate 97.7% detection outcomes. Consequently, the integration of the ECA module significantly contributes to the robustness and efficiency of the feature refinement process in medical image analysis.

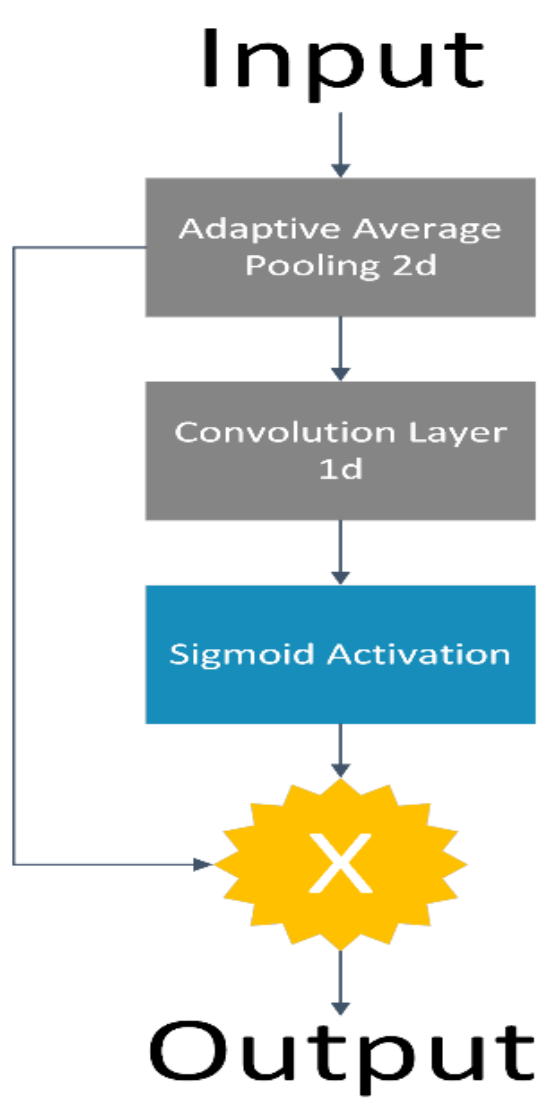

**Figure 6** Efficient Channel Attention

### C. Cross-Stage Partial (CSP) Fusion

To address the problem of duplicated gradient information in deep networks and to enrich the gradient flow, we incorporate Cross-Stage Partial (CSP) networks shown in figure 7 [48]. The CSP module splits the feature map into two parts. One part undergoes a dense block (a sequence of convolutions), while the other part bypasses it. The two paths are then concatenated. This design achieves two goals while maintaining optimal computational power. First, it reduces computational costs by only processing a fraction of the feature map through the expensive dense block. Second, it combines a non-processed feature map with a processed one, creating a richer gradient combination that improves learning ability and model accuracy.

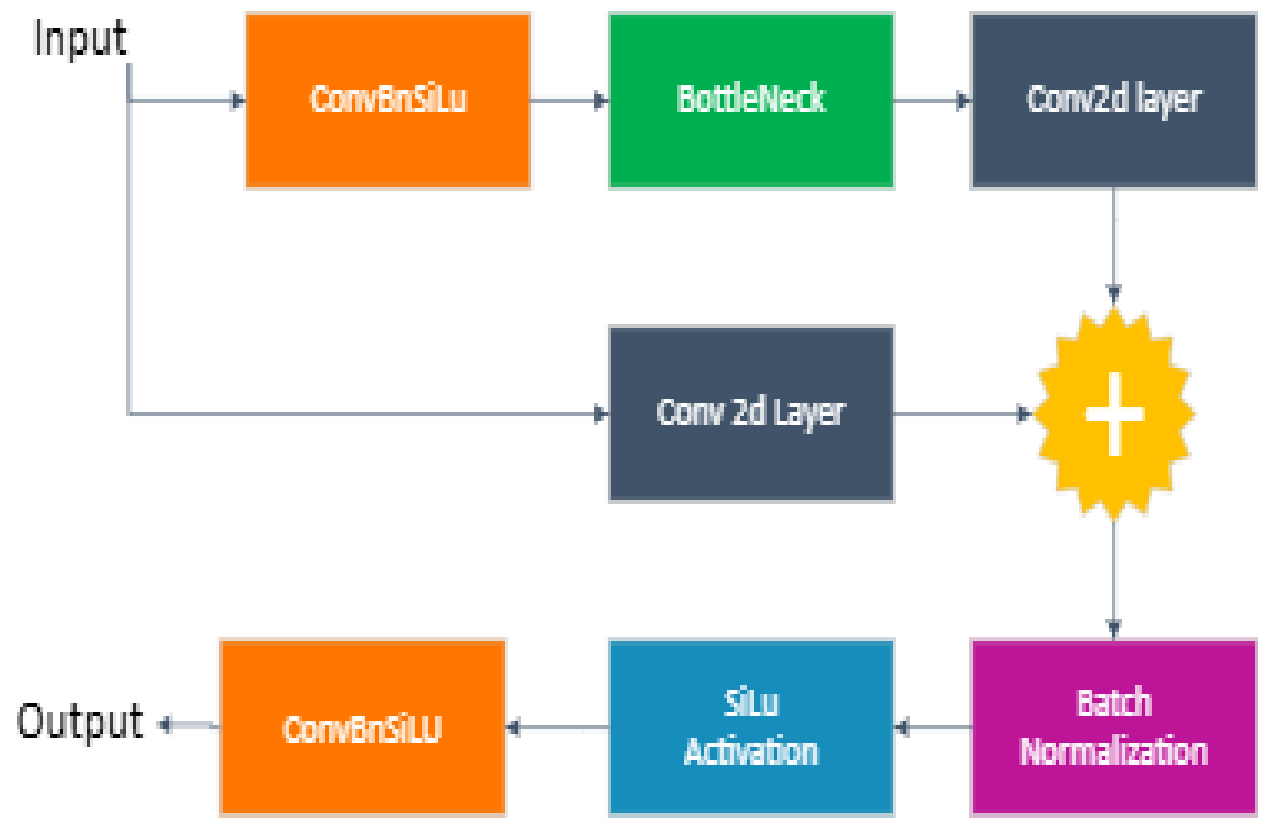

***Figure 7** Cross-Stage Partial (CSP) Fusion*

### 3.4.1 Neck: Path Aggregation Feature Pyramid Network (PA-FPN)

The Neck of our architecture is responsible for effectively aggregating multi-scale features extracted by the backbone. We employ an enhanced Path Aggregation Feature Pyramid Network (PA-FPN) to mitigate the effect of height variation and feature sizes in the imagery. The network-based division in mentioned in figure 9 where neck is meant for feature extraction from shallower to deeper feature maps. The standard FPN Rehman et al. [46] constructs a top-down pathway to combine high-level semantic features from deeper layers with high-resolution features from shallower layers. Our PA-FPN enhances this by two approaches. The first approach is Bottom-Up Path Augmentation**.** This adds a second bottom-up pathway from the lower layers to reduce the information path between lower and top-level features, making the localization signals for deeper layers stronger. The second approach is Adaptive Feature Pooling. It allows each proposal to access features from all levels of the feature pyramid, rather than being limited to a single level based on its size.

### 3.4.2 Head: Anchor-Free Decoupled Head

The Head is the component that makes the final predictions for disease detection. We adopted a modern Anchor-Free approach with a decoupled head structure.

A. **Anchor-Free (Point-based) Detection:** Traditional YOLO models use anchor boxes (pre-defined bounding boxes of various sizes and ratios) as detection ground truths. This can be sensitive to the choice of anchor sizes and adds complexity. Our anchor-free approach simplifies this by having the model predict the object's key point (usually the center) and its distance to the bounding box boundaries. This leads to a simpler, more efficient detection process that generalizes better to objects of novel sizes, such as the highly variable shapes of disease lesions in cotton leaves.

B. **Decoupled Head:** In a classic YOLO head, a single convolutional layer is tasked with predicting both the class (classification) and the bounding box coordinates (regression). This can create conflict during training. We use a decoupled head, which employs two separate small convolutional networks: one dedicated to classification and one dedicated to regression. This separation of both layers has been shown to significantly improve convergence and overall detection performance.

In the below modified YOLO Architecture, attention mechanisms are used in the feature fusion stage for dynamic feature selection and integration, allowing the model to automatically learn the importance of global and local information for adaptive integration. In the decoding stage, Deformable Convolutional Networks (DCN) are applied for up sampling to create the feature map of disease detection. This helps capture variations in disease lesion shapes, preserving fine boundary details in segmentation results.

Final CBAM (Convolutional Block Attention Module) is utilized in the proposed Cotton-Vision framework within the feature fusion stage of the model's head component. Although the main architectural innovation uses Efficient Channel Attention (ECA) in the backbone for lightweight feature recalibration, a final attention mechanism, referred to as CBAM, is applied per-scale to the FPN tensor before the final detection layers. This module integrates Channel Attention and Spatial Attention to achieve dynamic feature selection and integration for detecting edges, contours and boundary features. Its purpose is to automatically learn the importance of global and local information, refining the multi-scale features to emphasize subtle disease symptoms as in the case of CLCuV. The features are not visible to naked eye and can be confusing for other cotton diseases. This focused feature processing is crucial for the model's superior performance, contributing to a high precision of 97.7% in cotton disease detection.

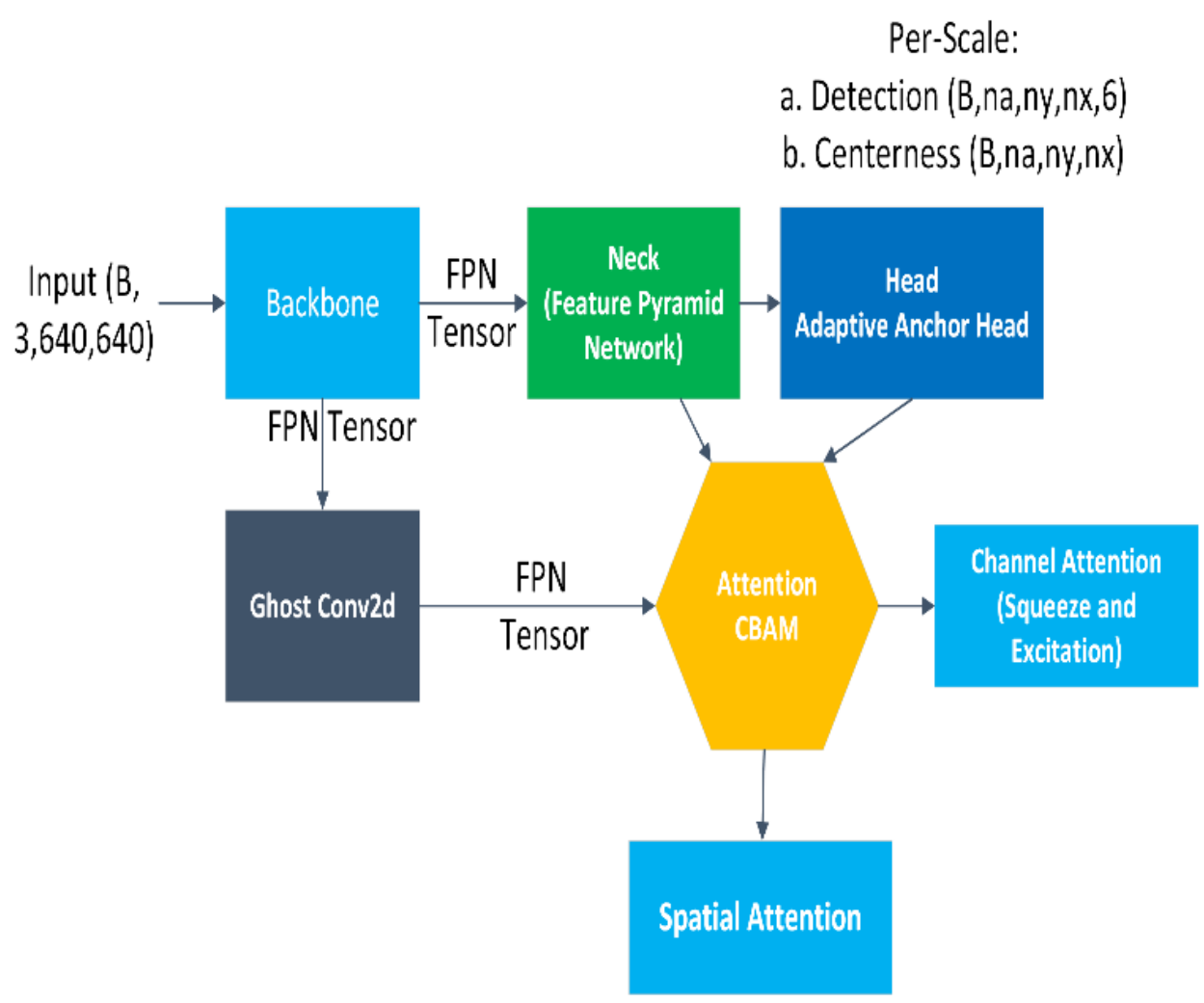

**Figure 8** CBAM Integration in the proposed Model architecture

## 3.5 Training Strategy

The model was trained using Stochastic Gradient Descent (SGD) with a momentum of 0.937 and weight decay of $5\times10^{-4}$. We employed a cosine annealing learning rate scheduler, decaying the initial rate of 0.01 down to 0.0001. A warm-up period of 3 epochs was used to stabilize training mixed precision training was utilized to accelerate training and reduce GPU memory consumption without sacrificing performance.

### 3.5.1 Data Augmentation

Achieving model robustness and generalization against the unpredictable elements of real-world UAV operations necessitated an extensive real-time data augmentation strategy. This comprehensive approach was essential to simulate the vast array of lighting, weather, and occlusion conditions the system would encounter in the field. The augmentation suite began with the mandatory application of the Mosaic image combination technique with probability=1.0. This was complemented by geometric and color transformations, including horizontal flipping (0.5), translation (±0.1), scaling (±0.5), and HSV-Hue augmentation (±0.015), which accounted for basic visual variance. Furthermore, we integrated advanced techniques like fog simulation and random erasing to directly prepare the model for challenging adverse conditions. This diverse strategy collectively ensures the final model maintains high predictive fidelity and stability across all encountered inputs.

### 3.5.2 Feature Extraction and Fusion by the Proposed Architecture

Cotton-Vision's feature, shown in **figure 9,** extraction and fusion strategy is crucial for detecting disease symptoms. A hierarchical feature extraction backbone captures features at multiple semantic levels. This ensures that the model captures both fine-grained and high-level information, improving against variable symptoms and field conditions [49]. Its efficiency and accuracy stem from three interconnected components.

First, the Backbone performs hierarchical feature extraction, capturing both fine-grained, high-resolution leaf details from early layers and crucial high-level semantic context regarding disease patterns from deeper layers. This backbone is optimized for speed and capacity using light-weight innovations like Ghost Bottlenecks and Efficient Channel Attention (ECA) modules.

Second, the Neck utilizes an enhanced Path Aggregation Feature Pyramid Network (PA-FPN) to resolve the issue of scale variance. It employs a top-down pathway to enrich spatial features with strong semantics and a bottom-up pathway to effectively preserve and propagate precise localization signals from shallow layers, ensuring accurate bounding box placement.

This multi-stage fusion process culminates in the Head, which outputs three balanced Multi-Scale Feature Maps (P3, P4, P5). These maps enable the decoupled detection heads to perform tailored predictions: P3 handles large, close-up objects; P4 covers medium-sized lesions; and P5 leverages strong semantic context to accurately identify small, distant, or early-stage disease manifestations. This comprehensive design guarantees robust and high-fidelity detection across the diverse conditions of real-world UAV operations.

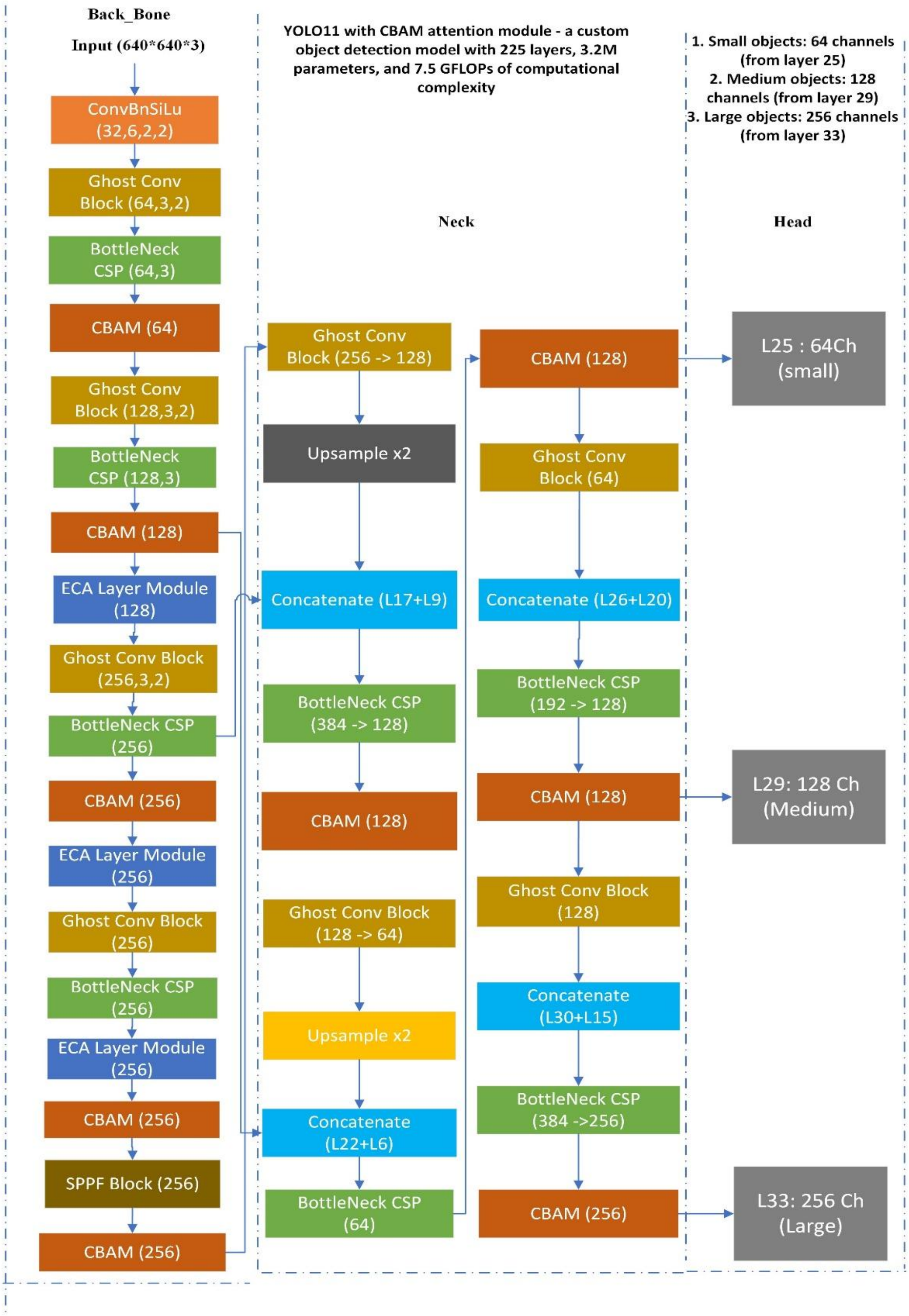

***Figure 9*** *Proposed architectural changes for Cotton-vision Model inspired from [46]architecture*

# 4. Experimental Setup

A reproducible experimental setup is fundamental to validating the performance of the Cotton-Vision model. This section details the software and hardware environment, the specific training parameters and strategies employed, and the comprehensive set of metrics used for evaluation.

## 4.1 Software and Hardware Environment

To ensure accessibility, reproducibility, and transparency, all development and experimentation were conducted on widely available, open-source platforms.

a) **Hardware:** The model was trained and evaluated using NVIDIA Tesla T4 GPUs, accessed via Google Colab and Kaggle platforms. The T4 GPU, with its 16 GB of VRAM, provided sufficient memory for batch processing and handling the high-resolution UAV imagery. The use of these cloud-based platforms guarantees that our results can be easily reproduced by other researchers without the need for specialized, proprietary hardware.

b) **Software:** The core implementation was carried out in Python 3.11.13. The deep learning framework of choice was PyTorch 2.6.0+cu12.4 and its companion library. This framework was selected for its flexibility, dynamic computation graph, and strong ecosystem for computer vision research. Key libraries for data manipulation were Pandas, numpy and image processing (opencv-python), and visualization (matplotlib, seaborn) were integral to the workflow.

## 4.2 Training Parameters and Strategy

The training process was carefully designed to maximize model performance while mitigating overfitting on the limited agricultural dataset.

### 4.2.1 Hyperparameters:

The training process was carefully tuned through a combination of hyperparameter optimization and robust augmentation strategies. The model was trained for 100 epochs with a batch size of 16, using AdamW optimizer with a momentum of 0.937 and weight decay of 0.0005. A cosine annealing learning rate schedule was adopted, starting at 0.01 and gradually reduced to 0.001, alongside a three-epoch warm-up phase for stability. To improve generalization, extensive data augmentation techniques including mosaic transformations, color-space modifications, and random scaling were applied. These strategies simulated diverse environmental conditions, such as varying lighting and occlusion, thereby enhancing robustness. Mixed-precision training further improved computational efficiency by reducing GPU memory requirements without compromising accuracy.

## 4.3 Disease Detection and Classification Metrics:

Evaluating computational efficiency is critical for UAV-based systems that often run on embedded hardware with constrained resources. The proposed model maintains a compact architecture with only 2.58 million parameters and approximately 6.3 GFLOPs per forward pass, making it significantly lighter than transformer-based baselines. Inference time averaged a few milliseconds per image on an NVIDIA Tesla T4 GPU. These results highlight that the integration of lightweight modules such as Ghost bottlenecks and ECA attention not only improved accuracy but also kept computational demands low. This balance between detection accuracy and efficiency is essential for ensuring scalability and deployment in real-world agricultural environments.

# 5. Results and Evaluations

This section presents a comprehensive evaluation of the Cotton-Vision framework, detailing its performance through quantitative metrics and a qualitative analysis. Cotton-Vision consistently outperformed competing models across all reported metrics. The Precision-Recall (PR) curves for all evaluated models are presented in **Figure 10**. The curve for Cotton-Vision encloses the largest area, corresponding to its highest mAP@0.5 score of 85.6%. This indicates a consistently superior trade-off between precision and recall across all confidence thresholds. The steeper initial decline in the curves of other models, such as YOLOv11n and RT-DETR, reflects their lower precision at higher recall rates, meaning they generate more false positives as the model becomes more sensitive.

This curve shows mAP evolution during training and validation across different Intersection over Union (IoU) thresholds. Rapid convergence close to 0.95 after about 40 epochs suggests that the model learned to detect and localize diseases accurately with minimal overfitting. Parallel decreases in training and validation loss curves further confirm stable learning dynamics and good generalization. The sustained high precision of Cotton-Vision across all recall levels is a testament to the effectiveness of its architectural enhancements in reducing misclassifications. The training loss indicates that the model is effective in training with losses decreasing to the point of loss without overfitting.

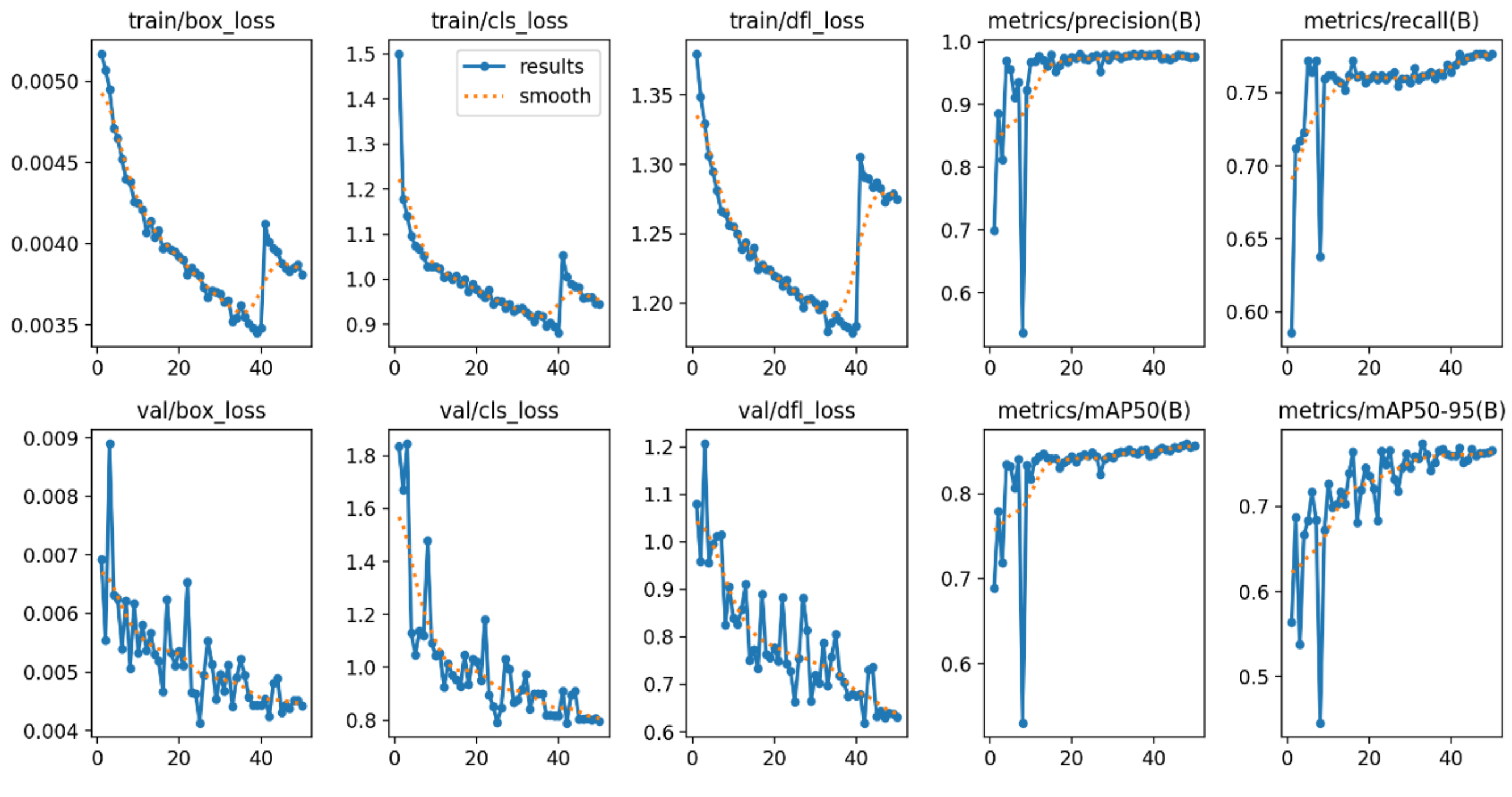

***Figure 10*** *Performance Evaluation graphical Representation for training of Cotton vision Model*

The performance evaluation further indicates that the Mean Average Precision (mAP) curves rapidly convergence, with the validation mAP (blue) closely tracking the training mAP (orange), indicating good generalization. The Loss curves decrease steeply and smoothly, suggesting effective learning without major instability. The validation loss (blue) consistently stays near or below the training loss (orange) across all metrics, which is an ideal result and indicates the model is not overfitting. This steady parallel decrease confirms the model's robustness, due to the architectural innovations i.e. the Ghost Bottleneck and ECA module mentioned in the paper. **Figure 11** qualitatively demonstrates the high performance of the Cotton-Vision model in detecting Cotton Leaf Curl Virus (CLCuV) from UAV imagery. The top-left example shows accurate detection of multiple disease instances (red bounding boxes) on cotton plants, even under challenging field conditions with varying scale and illumination. The model correctly classifies the affected areas with a high confidence score (0.84 and 0.90), suggesting strong visual feature learning. The bottom-right image, which appears to be a healthy field, shows no false positive detections, indicating excellent background suppression. The model effectively distinguishes the target disease from complex agricultural backgrounds, validating the efficacy of the architectural innovations like the Ghost Bottleneck and ECA modules for precision agriculture.

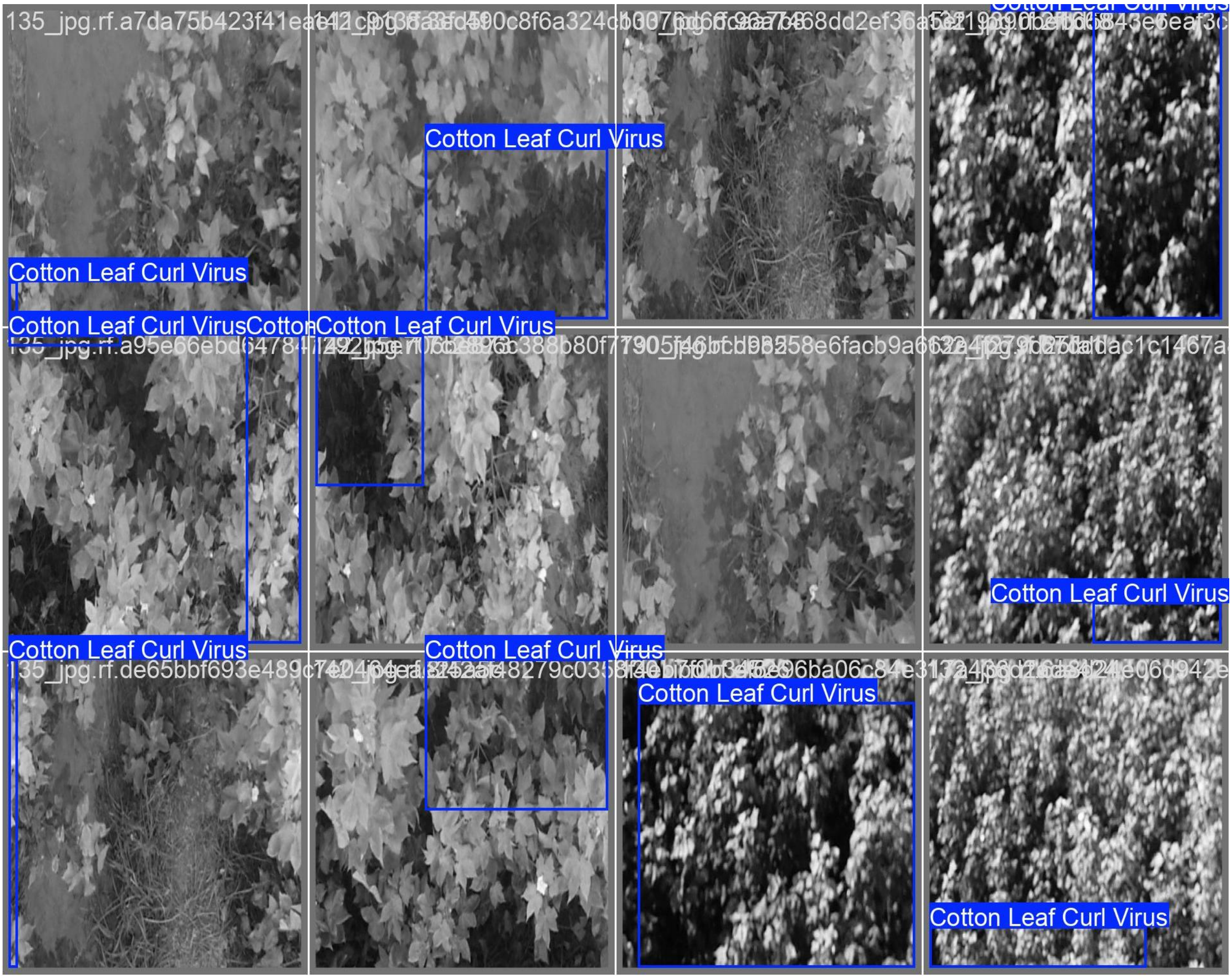

**Figure 11** Model Test Data output

As observed, YOLOv11n, while capable, exhibits several false negatives, failing to detect smaller or early-stage disease patches with lower contrast against the leaf background (e.g. second column). In contrast, our Cotton-Vision model demonstrates robust performance. It localizes disease symptoms with high confidence and 97.7% accuracy, minimizes false positives due to the ECA module's refined feature focus, and successfully identifies challenging, small-scale lesions thanks to the enhanced gradient flow from the CSP and Ghost modules. This visual analysis confirms the quantitative findings, illustrating that Cotton-Vision provides more reliable and precise detections suitable for practical agricultural scouting. Upon completion of the training, the framework achieved a precision of 97.7%, a recall of 77.7%, mAP@50 of 85.6%, and mAP@50–95 of 76.6%, surpassing the closest competitor by more than 20 percentage points in the most stringent IoU-based measure. These results highlight the model's ability to not only detect disease symptoms with high confidence but also localize disease with superior accuracy across varying spatial overlaps. When benchmarked against leading YOLO variants (v5, v8, v11, v12) and transformer-based detectors, Cotton-Vision achieved consistently superior results across all metrics. YOLO-based baselines often suffered from reduced recall and sensitivity to small lesion sizes, while transformer models showed improved recall but produced false positives due to misclassification of leaf textures and shadows. Cotton-Vision balanced both dimensions, yielding the most reliable overall performance. This comparative robustness confirms that the lightweight and attention-driven design enables Cotton-Vision to generalize better across the heterogeneous and noisy backgrounds common in UAV imagery.

**Table 3** Quantitative Comparison of different scores with existing YOLO family Architectures

| Model Name | Precision | Recall | mAP@0.5 | mAP@0.5:0.95 | Params (M) | GFLOPs |
|---|---|---|---|---|---|---|
| **YOLOv5n** | 90.7 | 59.4 | 64.4 | 43.0 | 2.654 | 7.8 |
| **YOLOv8n** | 70.0 | 60.0 | 62.0 | 33.0 | 3.157 | 8.9 |
| **YOLOv11n** | 91.1 | 60.8 | 68.7 | 48.1 | 2.590 | 6.4 |
| **YOLOv12n** | 88.2 | 58.8 | 63.4 | 44.7 | 2.55 | 6.2 |
| **YOLO-BenchMark** [46] | 88.3 | 66.2 | 75.1 | 50.7 | 3.99 | 6.3 |
| **Cotton**_Vision **(Ours)** | **97.7** | **77.7** | **85.6** | **76.6** | **3.19** | **7.2** |

The high precision is particularly significant in the agricultural context, as it minimizes false alarms and prevents unnecessary pesticide application. Meanwhile, the recall rate demonstrates that most diseased plants were successfully identified, striking a practical balance between sensitivity and specificity. The strong mAP@50–95 performance confirms robustness across a wide range of bounding box thresholds, an essential property for early-stage detection when lesions are small and poorly defined in the distant UAV Imagery with variation in detection.

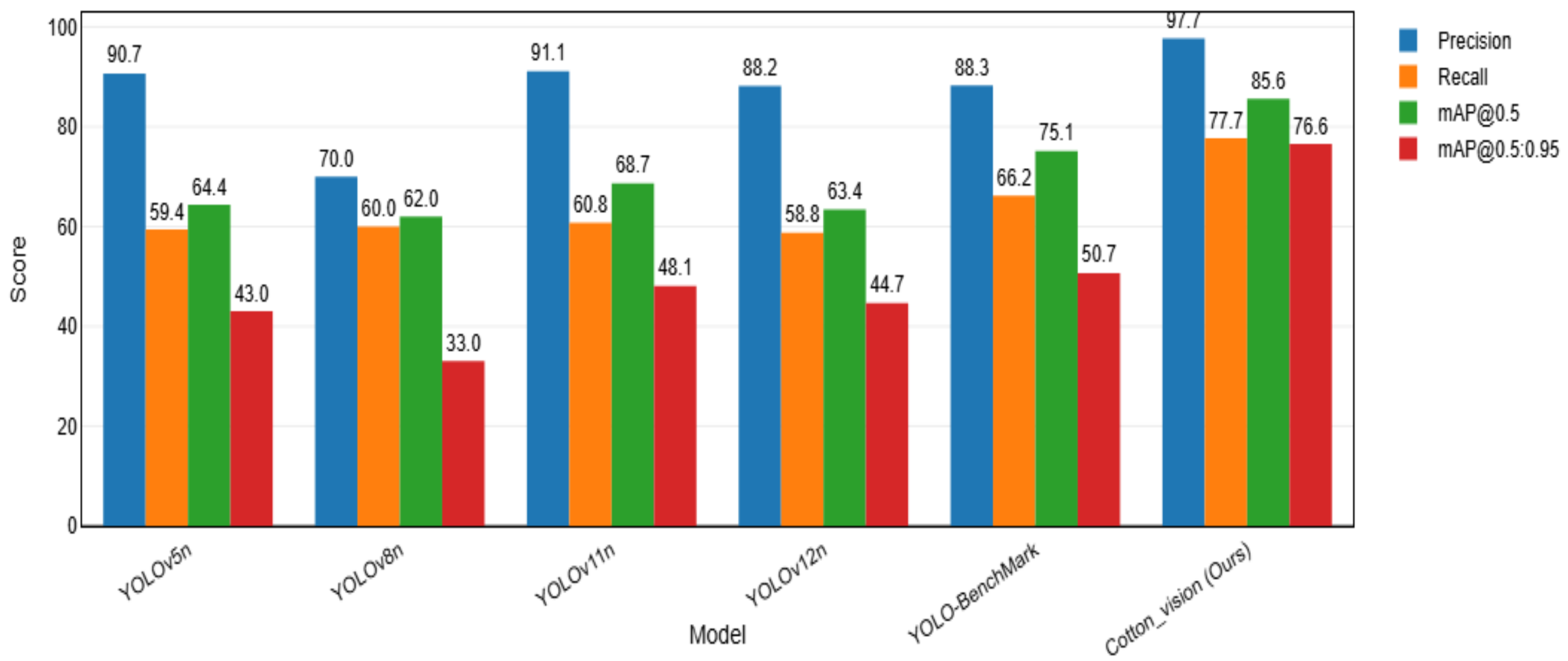

**Figure 12** Comparative Analysis of latest yolo models and their performance based upon provided evaluation matrices

Equally important, Cotton-Vision accomplishes these results with 2.58M parameters and 6.3 GFLOPs, markedly lighter than transformer-based detectors such as RT-DETR (32 GFLOPs). This computational efficiency indicates suitability for UAV-based, near real-time inference, a key requirement for scalable precision agriculture.

**Table 4** Quantitative comparison with transformer-based models

| Model Name | Precision | Recall | mAP:0.5 | mAP:05-0.95 | GFLOPS |
|---|---|---|---|---|---|
| **Cotton_Vision (Ours)** | **97.7** | **77.7** | **85.6** | **76.6** | 2.582 |
| **RT-DETR(l)-R50** | 82.9 | 61.6 | 59.9 | 50.3 | 32 |
| **RF-DTr (n)** | 84.7 | 64.0 | 69.8 | 51.7 | 29 |

To benchmark the performance of our Cotton-Vision framework, we conducted a comparative analysis against several state-of-the-art object detection models, including RT-DETR and RF-DTr, which represent transformer-based and lightweight YOLO-based architectures, respectively. As summarized in **Table 4**, our proposed model significantly surpasses the performance of these competitors across key metrics. Similarly, **Figure 13** indicates the superior performance of Cotton Vision on Sanjrani_CottonDB due to higher adaptability of feature learning and maintaining overall scores greater than RT_DTR and RF_DTR.

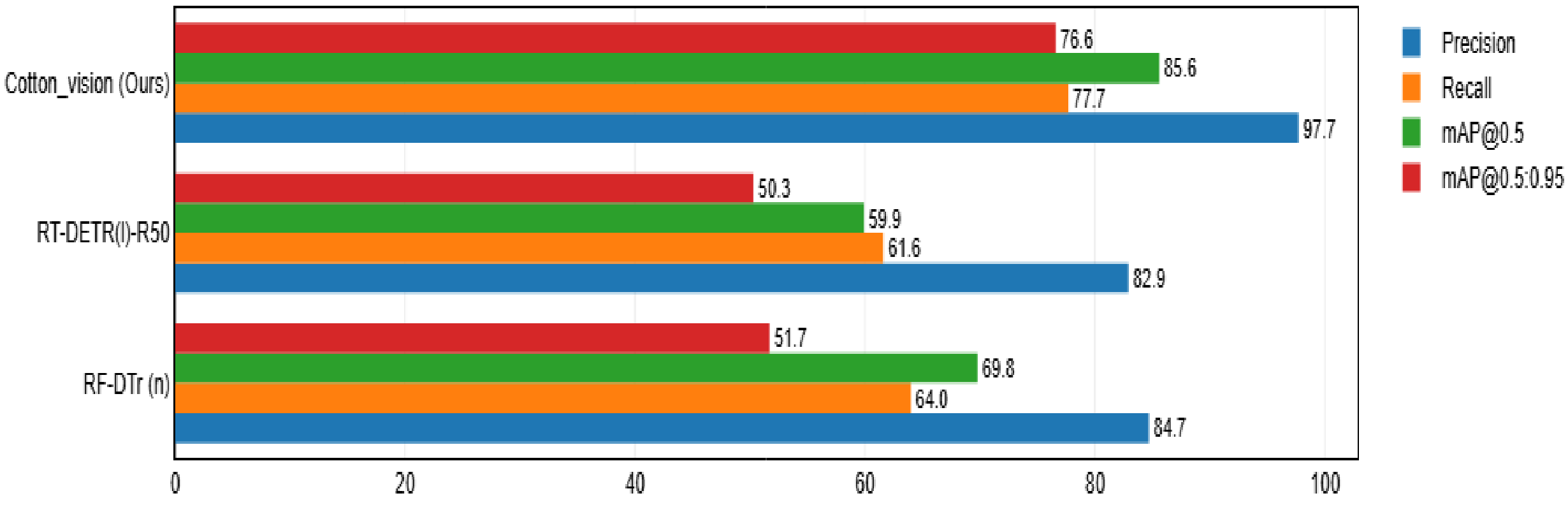

**Figure 13** Performance of transformer-based model vs Cotton vision model.

# 6. Ablation Study

The ablation study confirms the synergistic effect of our proposed modules. The addition of the Ghost Bottleneck modules alone resulted in a 3.4% improvement in mAP@0.5 while simultaneously reducing the model's parameters by a significant margin. Adding the ECA module further improved performance by 2.3%, demonstrating the value of adaptive feature recalibration. Finally, the integration of CSP fusion provided the most substantial performance boost, increasing the mAP@0.5 score by an impressive 15.5%, highlighting its critical role in enriching gradient flow and feature representation. The final model achieves its peak performance with the lowest parameter count, proving that our enhancements are not only effective but also highly efficient.

**Table 5** Ablation study table with different configuration tested during the training

| Configuration | mAP@0.5 | Params | ΔmAP |
|---|---|---|---|
| **Baseline (YOLOv11n)** | 68.7 | 2.590 | — |
| **+ Ghost** | 81.4 | 2.582 M | +12.7 |
| **+ ECA** | 84.9 | 2.64 M | +16.2 |
| **+ CSP** | 85.1 | 2.583 M | +16.4 |

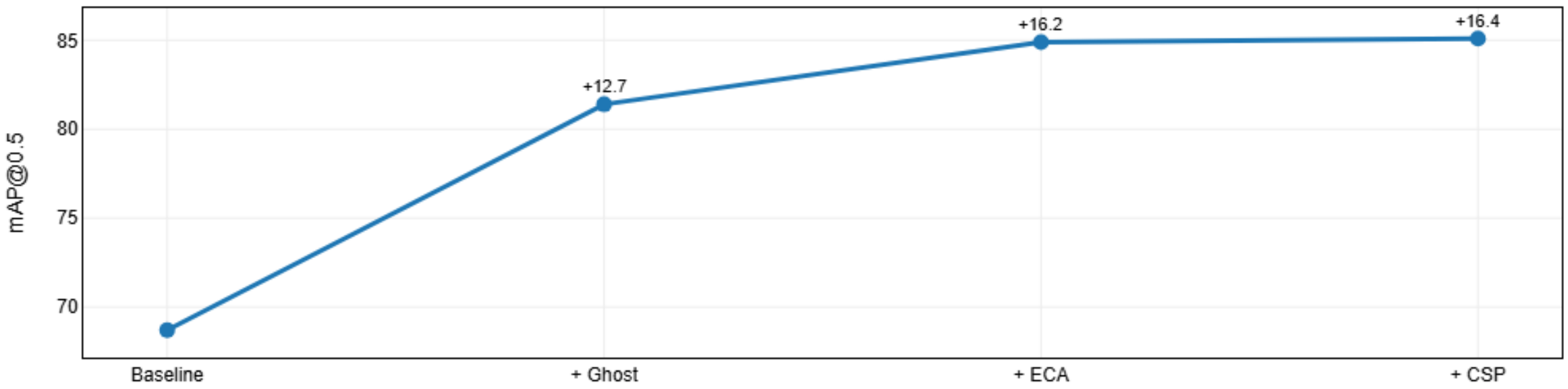

***Figure 14*** *Graphical representation of incremental increase in mAP in the ablation configuration.*

The addition of the Ghost bottleneck module led to a significant increase in performance, boosting the mAP@0.5 to 81.4%, a substantial gain of 12.7 percentage points, while simultaneously reducing the parameter count to 2.582 million. This demonstrates the Ghost bottleneck's effectiveness in generating rich feature maps with minimal computational overhead. The subsequent integration of the ECA module further enhanced the model's ability to focus on salient features, resulting in an mAP@0.5 of 84.9% (+3.5% gain). Finally, the inclusion of CSP fusion provided a marginal but valuable improvement, raising the mAP@0.5 to a final value of 85.1%. This systematic improvement at each step validates our design choices and demonstrates the synergistic effect of these lightweight modules.

# 7. Qualitative Analysis:

To complement the quantitative metrics, a qualitative assessment provides intuitive insight into the model's performance in real-world scenarios. **Figure 15** presents a comparative visualization of detection results of transformer-based models on sample images from the test dataset. The performance comparison between ground truth and detection results show close localization of diseased areas. The bounding box represents effected area for application of pesticides and other actionable insights.

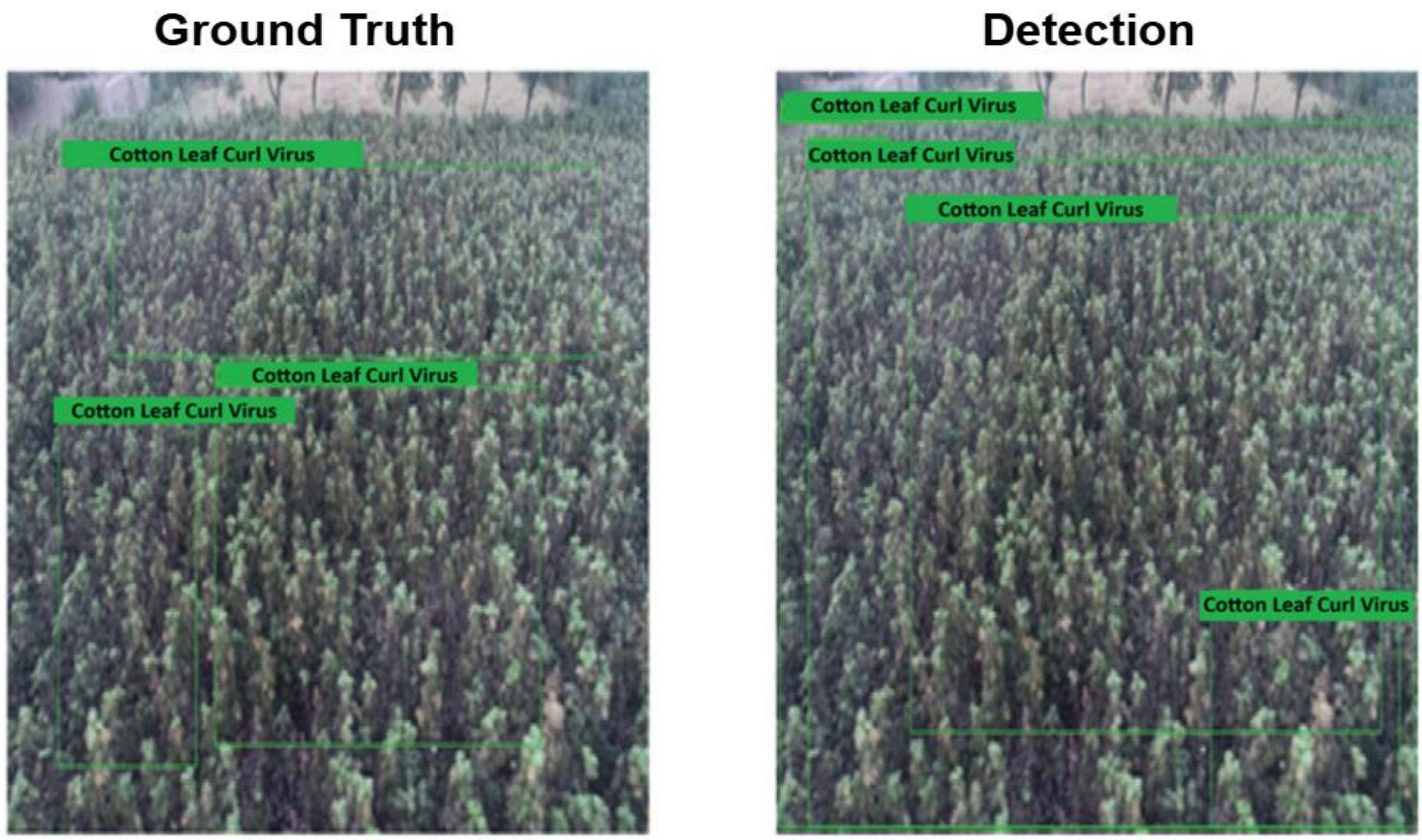

**Figure 15** Qualitative Analysis of Cotton_Vision model with ground truthing. The left-hand side of the results shows the expert annotation while right hand side is detection of the model on cotton UAV Imagery.

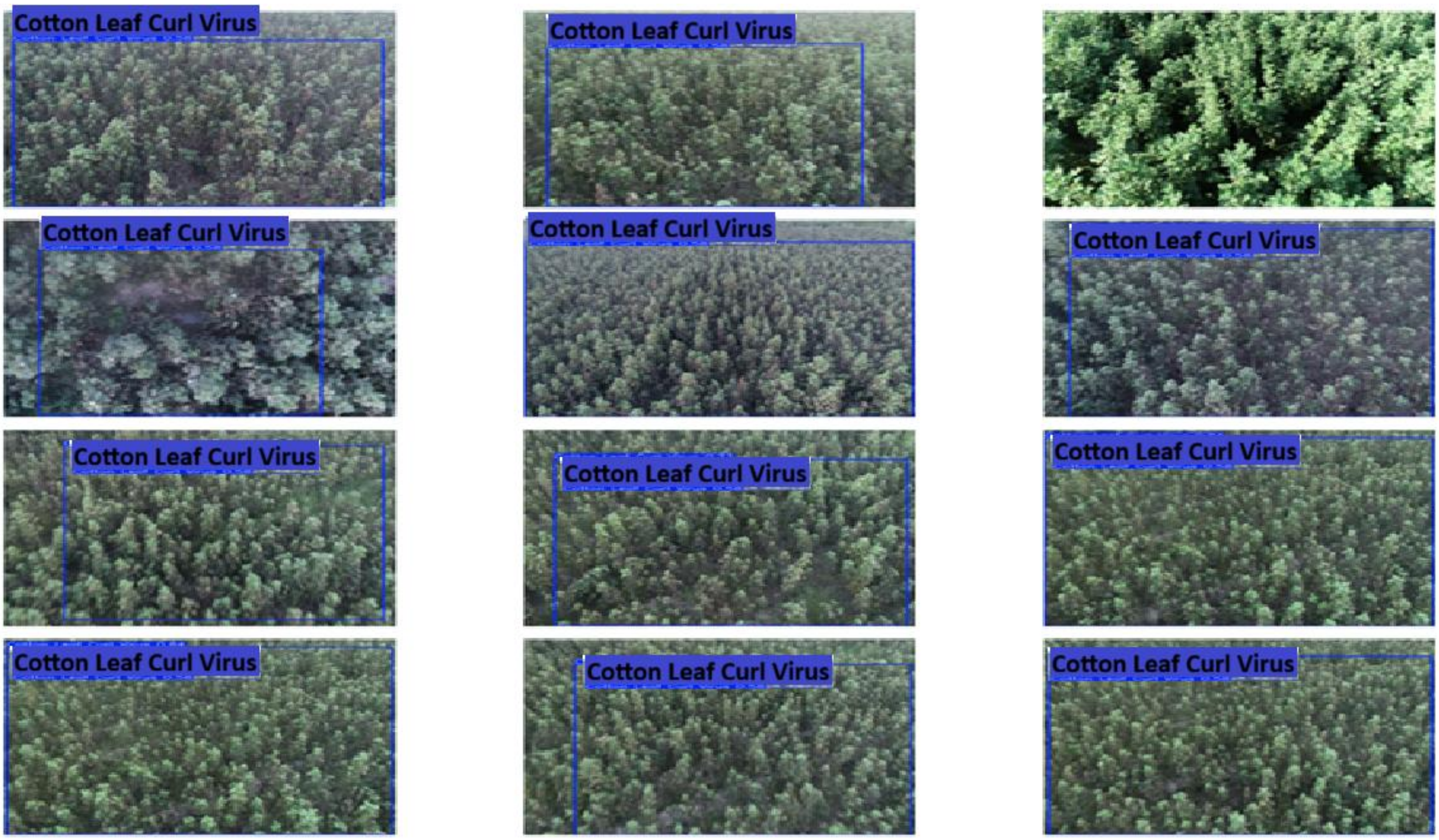

***Figure 16*** *Cotton Vision detection on Test dataset of UAV Imagery. The areas with the disease can be clearly seen with blue Bounding Box. The plants with healthy leaves show no detection of the disease.*

**Figure 16** exhibits the qualitative analysis supports the findings of transformer-based module. Visual inspection of detection maps shows that Cotton-Vision accurately identifies subtle, early-stage lesions that competing models often overlook. Unlike other models, which frequently misclassified shadow regions as diseased regions, Cotton-Vision maintained high confidence in true disease patches while minimizing false positives. The area detected in the bounding box suggests highest probability of disease area. This precise localization of disease provides a way forward for spot spraying and efficient application of pesticides for smart and sustainable agriculture practices.

# 8. Conclusion

This study introduced a lightweight, yet powerful deep learning framework designed specifically for UAV-based cotton disease detection. By incorporating modules such as Ghost Bottlenecks for parameter reduction, Efficient Channel Attention (ECA) for adaptive recalibration, and Cross-Stage Partial (CSP) fusion for enriched gradient flow, the proposed system achieved both high accuracy and computational efficiency. The integration of a Path Aggregation Feature Pyramid Network (PA-FPN) in the neck and an anchor-free decoupled head further enhanced the system's ability to capture subtle disease symptoms at multiple scales and make precise, reliable predictions. Extensive evaluations on the Sanjrani_CottonDB dataset demonstrated significant improvements over widely adopted models like YOLOv5, YOLOv8, and transformer-based architectures, achieving superior precision, recall, and mAP scores while maintaining low computational complexity. Extensive evaluation on the novel Sanjrani_CottonDB dataset demonstrates that Cotton-Vision outperforms a suite of leading models, including YOLOv5/v8/v11/v12 and RT-DETR, achieving a precision of 97.7% and a mAP@0.5 of 85.6%. The ablation study conclusively proves the improvement in accuracy of each architectural component to this superior performance. This work provides an accurate technological solution that moves precision agriculture for cotton closer to practical, scalable solution for early disease detection in cotton fields.

# 9. Future Work

Although the presented framework demonstrates strong accuracy and efficiency, several avenues remain open for further improvement and exploration. One important direction is multimodal sensor fusion, where RGB imagery from UAVs could be combined with multispectral, hyperspectral, or thermal data to capture physiological changes in crops that are invisible to the naked eye [23],[50]. This would allow disease detection at earlier, pre-symptomatic stages and provide a richer understanding of plant health. Another promising area is domain adaptation and federated learning, which could help overcome the limitation of dataset generalizability [51]. By enabling models to learn from diverse datasets collected across different regions and climates without centralizing data, these approaches can improve robustness while maintaining privacy.

Techniques like quantization, pruning, and Tensor RT optimization could be leveraged for this purpose [52]. In addition, multi-task learning represents another promising extension, where the system could simultaneously detect diseases, estimate severity levels, and even perform plant counting, offering farmers a more comprehensive decision-support tool. Finally, broader adoption will require interdisciplinary work beyond model development. Collaboration with agronomists, engineers, and local farming communities is necessary to translate technical advances into usable solutions that address socioeconomic and logistical barriers. This includes designing farmer-friendly interfaces, integrating detection systems into mobile platforms, and exploring cost-effective UAV solutions tailored to smallholder farmers. By pursuing these future directions, AI-driven detection systems can evolve from research prototypes into practical, field-ready technologies that significantly contribute to agricultural sustainability and resilience.

## Funding:

No Funding was acquired to carry out this research.

## Data Availability:

Data will be available from the corresponding author upon reasonable request.